\documentclass[letterpaper, 10 pt, conference]{ieeeconf}  

\IEEEoverridecommandlockouts                              

\usepackage{booktabs,balance,subfigure}
\usepackage{graphicx,balance}
\usepackage{multirow}
\usepackage[sort,compress]{cite}
\let\labelindent\relax
\usepackage{enumitem}
\usepackage{xcolor}
\usepackage{amsmath}
\usepackage{hyperref}
\hypersetup{
    colorlinks=true,
    urlcolor=black,
    linkcolor=black
    }
\setlist[itemize]{align=parleft,left=0pt..1em}

\title{\LARGE \bf 
Wide-Area Geolocalization with a Limited Field of View Camera
}

\author{Lena M. Downes$^{1, 2, 3}$, Ted J. Steiner$^{2}$, Rebecca L. Russell$^{2}$ and Jonathan P. How$^{1}$
\thanks{$^{1}$Department of Aeronautics and Astronautics,
        Massachusetts Institute of Technology, Cambridge, MA, USA}%
\thanks{$^{2}$ Perception and Embedded ML Group,
        Draper, Cambridge, MA, USA}%
\thanks{$^{3}$ Draper Scholar. Research funded by Draper. {\tt\small lmdownes@mit.edu}}%
}

\begin{document}

\maketitle
\thispagestyle{empty}
\pagestyle{empty}

\begin{abstract}
Cross-view geolocalization, a supplement or replacement for GPS, localizes an agent within a search area by matching images taken from a ground-view camera to overhead images taken from satellites or aircraft. Although the viewpoint disparity between ground and overhead images makes cross-view geolocalization challenging, significant progress has been made assuming that the ground agent has access to a panoramic camera. For example, our prior work (WAG) introduced changes in search area discretization, training loss, and particle filter weighting that enabled city-scale panoramic cross-view geolocalization. However, panoramic cameras are not widely used in existing robotic platforms due to their complexity and cost. Non-panoramic cross-view geolocalization is more applicable for robotics, but is also more challenging. This paper presents Restricted FOV Wide-Area Geolocalization (ReWAG), a cross-view geolocalization approach that generalizes WAG for use with standard, non-panoramic ground cameras by creating pose-aware embeddings and providing a strategy to incorporate particle pose into the Siamese network. ReWAG is a neural network and particle filter system that is able to globally localize a mobile agent in a GPS-denied environment with only odometry and a 90\textdegree{} FOV camera, achieving similar localization accuracy as what WAG achieved with a panoramic camera and improving localization accuracy by a factor of 100 compared to a baseline vision transformer (ViT) approach. 

\end{abstract}

\section{INTRODUCTION}
GPS is an external system for localization that is susceptible to failure through jamming, spoofing, and signal dropout due to dense foliage or urban canyons. Cross-view geolocalization \cite{Tian, Shi, Cai, Kim, Hu} is a localization method that only requires images from a ground-view camera and preexisting overhead imagery, with or without GPS measurements. Cross-view geolocalization measures the similarity between a ground image and all of the satellite images in a search area to determine the location that the ground image was taken from. Satellite imagery at some resolution is widely available for most of the planet, even in forests and urban areas where GPS signals can be weaker.
\begin{figure}[t!]
  \centering
  \includegraphics[width=\linewidth]{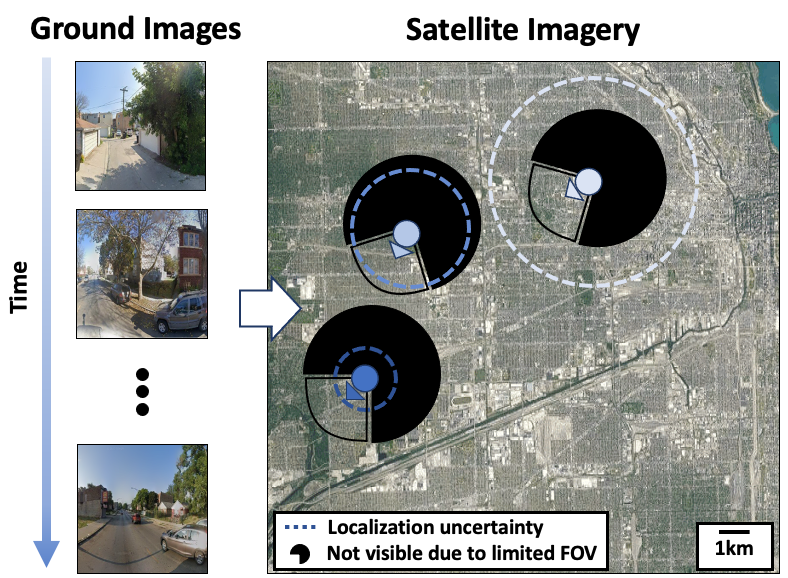}
    \vspace*{-0.25in}
  \caption{ReWAG is a cross-view geolocalization system that takes in a series of non-panoramic ground-view camera images and satellite imagery of the search area to accurately localize the agent on a search area scale that was not possible with previous works.}
  \label{fig:intro}\vspace*{-.3in}
\end{figure}

Cross-view geolocalization with ground and overhead images is challenging due to the wide difference in viewpoints. Many existing works \cite{Hu, ZhuVIGOR, Liu2019, Viswanathan} rely upon the use of a panoramic ground camera because it effectively decreases the problem dimensionality by reducing the impact of heading. Although a panoramic ground camera can have different headings, the heading only affects the alignment of the image, not the content, whereas the heading of a limited field of view (FOV) camera affects the visible content of the image. The use of panoramic ground cameras also simplifies the problem by maintaining as much semantic similarity as possible between the two viewpoints---overhead images show 360\textdegree{} of the surroundings of a ground agent, as do panoramic ground cameras. However, in practice panoramic cameras are rarely used due to their high monetary cost (resulting in lesser availability) and their difficulty to mount without occlusion. As a result, few real-world systems can benefit from panoramic-based localization. Widespread adoption of cross-view geolocalization technology will require its applicability to platforms without panoramic imaging capabilities, like that shown in Fig.~\ref{fig:intro}. 

Most recent work on cross-view geolocalization takes a deep learning approach to the problem by using Siamese networks \cite{Kim, Hu, Cai, Liu2019, Zhu2021, ZhuVIGOR}. A Siamese network consists of a pair of neural networks with matching architectures that simultaneously learn embedding schemes for ground and overhead images. The Siamese network is trained to embed images so that images that were taken in the same location are close together in embedding space while images that were taken in different locations are far apart in embedding space. Although Siamese networks have improved geolocalization performance beyond what was demonstrated with hand-crafted features, their accuracy can be improved further by integrating their measurements over time with a particle filter \cite{Xia, ZhuVIGOR, Hu, Kim, downes_iros_22}. However, existing particle filter geolocalization works are highly constrained--- requiring some level of GPS data \cite{Xia, ZhuVIGOR}, 180\textdegree{} or greater FOV of the ground camera \cite{Xia, ZhuVIGOR, Hu, Kim, downes_iros_22}, perfect initial location knowledge \cite{Hu}, or a search area of less than 2.5 km$^2$ \cite{Kim, Hu}. These works all impose additional constraints because they are not able to efficiently geolocalize a limited FOV camera across a large search area. Localization with a limited FOV camera increases both the difficulty of the problem and the computational requirements.

\textbf{Our contribution.} Our approach, Restricted FOV Wide-Area Geolocalization (ReWAG), builds upon our previous work in \cite{downes_iros_22} to enable efficient wide-area geolocalization with a restricted FOV camera through two key changes. The first change is a computationally efficient method for matching limited FOV ground images to satellite images by appending relative pose information to the intermediate network embedding before inputting it to the Spatial Aware Feature Aggregator (SAFA) \cite{Shi}. The second change is the dual incorporation of relative pose into both the Siamese network and the particle filter, which enables the probability distribution to be modeled more accurately. The only additional information these changes necessitate at runtime is a noisy heading from a compass, and they produce a cross-view geolocalization system that is capable of localizing across city-scale search areas using a 90\textdegree{} FOV camera. In summary, in this paper we demonstrate the following contributions of ReWAG for restricted FOV localization: 
\begin{enumerate}[leftmargin=*,topsep=-1em]
    \item Efficient pose-aware embedding generation,
    \item A particle filter system that more accurately models the probability distribution, resulting in lower average and final localization error, and
    \item Faster particle filter convergence than a ViT baseline \cite{zhu2022transgeo}.
\end{enumerate}

\section{RELATED WORKS}
\textbf{Ground-to-aerial cross-view geolocalization.} Cross-view geolocalization derives from previous work in the areas of scene recognition and image retrieval. Ground-to-aerial cross-view geolocalization pushes previous work to higher levels of difficulty due to the vast difference in viewpoints between ground and aerial images. Previous works have attempted to solve this problem with hand-crafted features and traditional computer vision techniques \cite{Viswanathan, viswanathan2016, jacobs, bansal}, but recent works \cite{Workman, Kim, Hu, Cai, Vo, Rodrigues, Shi, Cao, Zhu2021, Liu2019, lin, lin2015} have mostly applied deep learning in the form of Siamese networks \cite{bromley}. In recent years, recall at top-1 has been steadily rising, but most works \cite{Hu, ZhuVIGOR, Liu2019, Viswanathan} focus on panoramic ground images and report high recall at top-1 for these panoramic ground images. Recall at top-1 for limited FOV ground images is significantly lower than that for panoramic ground images. However, limited FOV cameras are much more common than panoramic cameras for robotic applications.

\textbf{Orientation-aware cross-view geolocalization.}
Non-panoramic ground cameras make cross-view geolocalization more difficult due to the reduced number of visible features in the image and due to the matching satellite features being concentrated within one area of the satellite image instead of spread throughout it. The unknown orientation of the ground camera is a key factor in this problem. Some previous works have developed methods to incorporate an understanding of orientation into the cross-view geolocalization system, like by appending orientation maps to images to be input to the Siamese network \cite{Liu2019}, by using Dynamic Similarity Matching (DSM) to calculate the correlation between ground and satellite images \cite{shi2020looking}, or by jointly embedding the full satellite image as well as the satellite image portion that is visible in the limited FOV ground image \cite{rodrigues2022global}. Instead of jointly determining the most highly matching satellite image and the orientation, \cite{shi2022beyond} assumes that the satellite image has already been determined, and they then use pose optimization to estimate the pose within that satellite image. \textit{These works, not designed for mobile robotics constraints, require many search iterations, polar transformations and data augmentations at runtime and hence may be too computationally demanding for real-time robotics.}

\textbf{Orientation-blind cross-view geolocalization.}
More recent works tend to treat orientation as an aspect of the problem that can be solved at the last step \cite{Zhu2021}, or do not directly encode or estimate it at all \cite{Shi, zhu2022transgeo}. In \cite{Zhu2021} orientation-invariant embedding schemes are learned through a combination of global mining, binomial loss, and training data augmentation with random rotations. Spatial Aware Feature Aggregation (SAFA) \cite{Shi} is an attention mechanism that helps the network to learn image descriptors regardless of the large viewpoint difference between ground and satellite views. TransGeo \cite{zhu2022transgeo} uses a vision transformer instead of the typical convolutional neural network (CNN) approach. This attention-focused approach embeds patches of the images into tokens with learnable position tokens, which gives it a more flexible method for learning about orientation and position while embedding images. \textit{Although orientation-blind embedding schemes improve image retrieval when orientation is unknown, these methods do not have a mechanism by which the ground camera pose can be input when modeling with a particle filter.}

\begin{figure*}[t!]
\centering
  \includegraphics[width=0.9\linewidth]{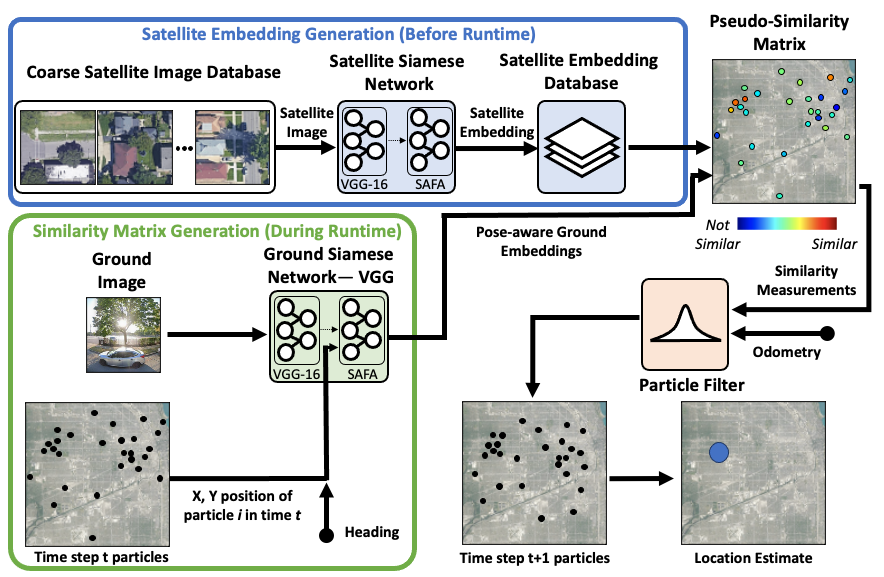}
  \caption{Diagram of ReWAG. Satellite embeddings are generated before runtime with the coarsely sampled satellite image database and the Siamese network that was trained with trinomial loss. During runtime, a pose-aware ground embedding is generated for each particle at each time step and combined with the satellite embeddings to create a pseudo-similarity matrix, which is the similarity of each particle with its location in the search area. Odometry and measurements from the pseudo-similarity matrix are input to the particle filter with a Gaussian measurement model to generate a location estimate.}
  \label{fig:system}
\vspace*{-0.2in}
\end{figure*}

\textbf{Particle filter implementations.} 
Some previous works have combined image retrieval with particle filters to enable cross-view geolocalization over time as an agent moves through a search area \cite{Kim,Hu,Xia, Viswanathan, viswanathan2016, downes_iros_22}. Particle filters in these systems use random discrete particles to model a probability distribution of the agent location given odometry, satellite images of the search area, and ground images. However, existing works have localized with particle filters that use wide angle ground cameras \cite{Hu, downes_iros_22, Viswanathan, viswanathan2016}.  Although \cite{Kim,Xia} localize a ground agent with a 180\textdegree{} ground camera instead of a panoramic 360\textdegree{} camera, commonly available cameras like those found in cell phones have FOV of 90\textdegree{} or less, making much fewer features visible in their images. \textit{These existing particle filter cross-view geolocalization systems are not able to accurately localize with extremely limited FOV images.}

\vspace{-0.1in}
\section{METHODS}
\subsection{Overview of Approach}
ReWAG builds upon WAG, the system developed by \cite{downes_iros_22}, to enable localization across a wide search area with restricted FOV ground images. ReWAG uses WAG's strategies of creating a coarse satellite image database, generating embeddings with Siamese networks based on VGG-16 \cite{vgg} and SAFA \cite{Shi}, and localizing over time with a particle filter (see Fig.~\ref{fig:system}). However, ReWAG differs from WAG in two major ways: first, ReWAG generates pose-aware image embeddings in a computationally efficient manner, and second, ReWAG generates more informative similarity measures and hence more accurately models the agent location probability distribution by incorporating pose information from each particle into the Siamese network input. For non-panoramic ground imagery it is necessary to incorporate this additional information due to the increased difficulty of the problem.

ReWAG first coarsely samples the search area to construct a database of satellite images that are preprocessed with a Siamese network before runtime to generate satellite embeddings. During runtime, a generic ground embedding is generated by the VGG-16 portion of the Siamese network for each ground image, and for each particle a pose-aware embedding is produced from the generic embedding and the particle's pose. The similarity between each particle's pose-aware ground embedding and its corresponding satellite embedding is calculated to produce the pseudo-similarity matrix, a probabilistic representation of the ground image's similarity across the search area. The particle filter receives odometry and measurements from the pseudo-similarity matrix to produce a location estimate at each time step. 

\subsection{Pose-Aware Embeddings}
\begin{figure}[t!]
\centering
  \includegraphics[width=\linewidth]{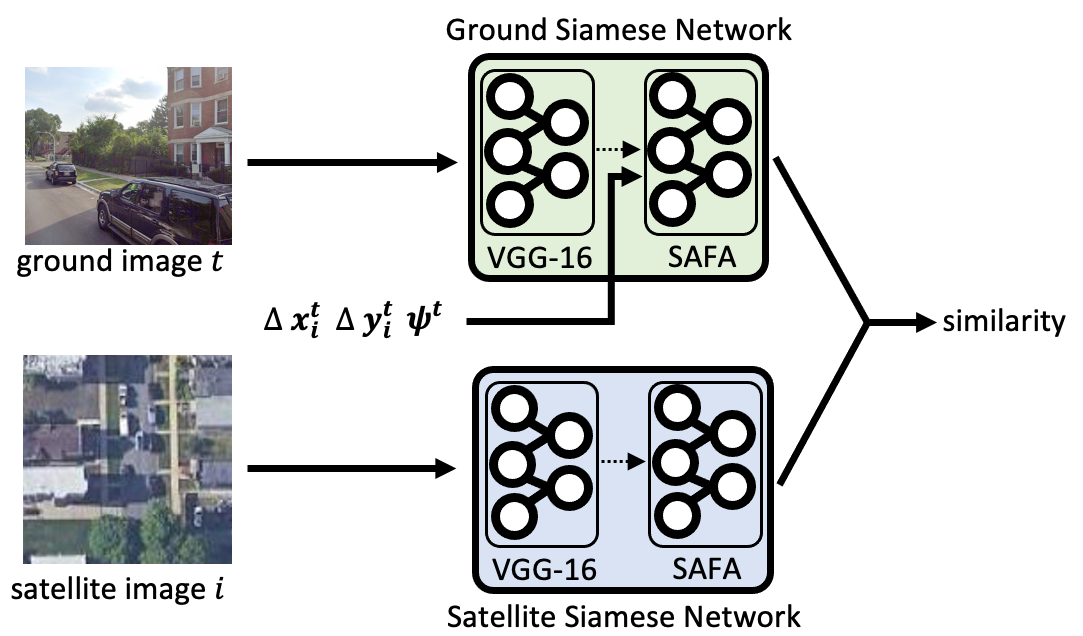}
  \vspace*{-0.3in}
  \caption{Our method requires pose to be incorporated in only the SAFA portion of the ground Siamese network, which reduces computation and allows faster inference at run time.}
  \label{fig:pose_aware}\vspace*{-0.218in}
\end{figure}

We have developed a method to train the ground Siamese network to generate pose-aware embeddings in a computationally efficient manner while minimally modifying the architecture. Like WAG, our Siamese network architecture is derived from that of \cite{ZhuVIGOR}, which consists of a VGG-16 backbone and a SAFA module to increase the network's spatial understanding. In ReWAG, the ground Siamese network is modified to append the particle pose to the intermediate embedding that is output by the VGG-16 backbone, as shown in Fig.~\ref{fig:pose_aware}. This intermediate embedding appended with the particle pose is then input to SAFA, which learns a spatial-aware representation of the ground image.  

ReWAG's computationally efficient benefit comes from the ability to generate one base embedding for each ground image, and then append any pose to the base embedding to efficiently generate a pose-aware embedding for each particle with the much lighter-weight SAFA. When combined with a particle filter, this design enables the VGG-16 inference to be done once per time step instead of once for each particle for each time step. In contrast, \cite{rodrigues2022global} generates pose-aware embeddings through a joint global and local pipeline, hence the full embedding generation must be done for each particle at each timestep.

\subsection{Siamese Network and Particle Filter Integration}
\begin{figure}[t!]
\centering
  \includegraphics[width=\linewidth]{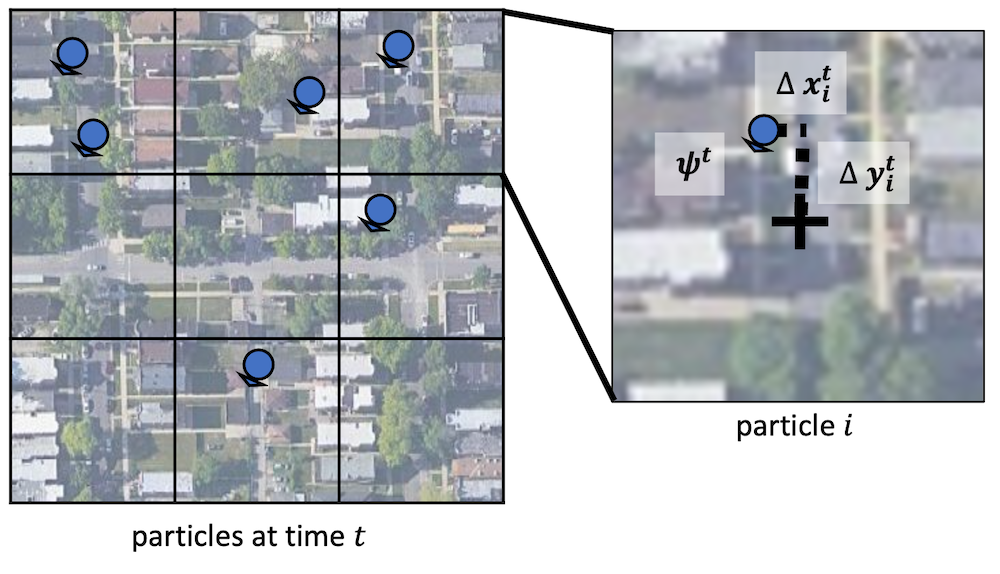}
  \caption{Particles are dispersed through search area, which is segmented into satellite tiles whose embeddings are precomputed before runtime. At each time step, the true heading of the ground agent is approximately known and the x, y location of each particle within its satellite tile is given by its displacement from the center of the tile. }
  \label{fig:particles}\vspace*{-0.2in}
\end{figure}

Our key observation is that cross-view geolocalization can be improved by more thorough integration between the Siamese network and the particle filter. Previous works have built systems with mostly one-way connections between the Siamese network and the particle filter---for each time step, the location of each particle determines which satellite image will be compared with that time step's ground image, and that similarity is used to adjust the weight of that particle. However, additional information can be integrated into the Siamese network-particle filter connection. In addition to a corresponding satellite image, each particle also has a location within that satellite image, as shown in Fig.~\ref{fig:particles}. the particle filter is modeling a probability distribution that includes the location within the satellite image, but traditional architectures do not factor that information into the similarity measure and hence it is not reflected in the particle weights. 

We have developed a method by which we incorporate particle pose information into the similarity measure through the pose-aware embeddings. The pose of a particle $i$ at time $t$ consists of x and y displacements $\Delta x^t_i$ and $\Delta y^t_i$, which are determined from the particle's location within its satellite tile, and the heading $\psi^t$, which is determined from sensor measurements at time step $t$. We assume that the heading measurement will be fairly accurate to the true ground agent heading based on the typical error of a compass. At each time step, the ground image is used to generate one generic intermediate embedding, and for each particle at that time step a pose-aware embedding is generated from the generic embedding and each particle pose. This method increases the computation required for each particle filter update, but the computationally efficient method by which we generate pose-aware embeddings helps to offset this increase. The incorporation of the particle pose aids the Siamese network in identifying where within the satellite image there should be corresponding features if the image pair is a positive match. Without pose-aware embeddings, a negative image pair could incorrectly find matching features anywhere within the satellite image. With pose-aware embeddings, a negative image pair will be encouraged to only look for matching features within a specified portion of the satellite image, decreasing the opportunity to find false matches.


\section{RESULTS}
\subsection{Experimental Setup}
To demonstrate ReWAG's performance relative to WAG \cite{downes_iros_22}, we perform limited FOV ground image localization experiments with the same test paths from \cite{downes_iros_22} but, instead of using panoramic ground images, we crop the ground images to 90\textdegree{} FOV. These test paths are a large scale localization experiment with very noisy location initialization across the entire city of Chicago using simulated data. The simulated data consists of ground images and overhead satellite images from Google Maps Static API and odometry measurements from the ground-truth displacement between images with added noise proportional to displacement. A satellite image database was generated by sampling the search area approximately every 60 meters into a 256 $\times$ 256 grid of non-overlapping satellite image tiles. This grid size maintains a similar image size and resolution that the network was trained on; satellite images of size 640 $\times$ 640 pixels with a resolution of approximately 0.1 m/pixel.  

We use a version of the neural network architecture from \cite{ZhuVIGOR} with the VGG-16 backbone and the SAFA module, modified to generate pose-aware embeddings. The satellite network architecture is unchanged from \cite{downes_iros_22}.  We train both ReWAG's Siamese network and our comparison baseline, a stage-1 TransGeo \cite{zhu2022transgeo}, on the VIGOR dataset \cite{ZhuVIGOR} with the ground images cropped to 90\textdegree{}. We train the TransGeo baseline for 50 epochs with the training parameters described in \cite{zhu2022transgeo}, and ReWAG for 30 epochs with triplet loss \cite{Hu}:
\begin{equation}\label{eq:triplet}
\mathcal{L}_{\text{triplet}} = \log\left(1+e^{-\alpha\left(d_{pos} - d_{neg}\right)}\right)
\end{equation}
with $\alpha$ loss parameter set to 10. Then we train ReWAG for 15 additional epochs with trinomial loss \cite{downes_iros_22}:
\begin{equation}\label{eq:trinomial}
\begin{aligned}
\mathcal{L}_{\text{t}} = \frac{\log\left(1+e^{-\alpha_{\text{p}}\left(S_{\text{p}}-m_{\text{p}}\right)}\right)}{N_{\text{p}}\alpha_{\text{p}}} +  \frac{\log\left(1+e^{\alpha_{\text{n}}\left(S_{\text{n}}-m_{\text{n}}\right)}\right)}{N_{\text{n}}\alpha_{\text{n}}} \\
+\frac{\log\left(1+e^{-\alpha_{\text{semi}}\left(S_{\text{semi}}-m_{\text{semi}}\right)}\right)}{N_{\text{semi}}\alpha_{\text{semi}}} 
\end{aligned}
\end{equation}

with the parameter values used in \cite{downes_iros_22}. The filter has 30,000 particles and uses the Gaussian measurement model from \cite{downes_iros_22}. We initialized the particle filter with Gaussian distributions, centered 1.3 km from the true initial location (standard deviation of 900 m) for C-1 and C-2, and centered 600 m from the true initial location (standard deviation of 300 m) for C-3. We add 2\% noise to the ground-truth odometry and 1\% noise to the ground-truth heading at each time step.



\subsection{Large-scale Test: Chicago} 
\begin{figure}[t!]
\centering
  \includegraphics[width=0.8\linewidth]{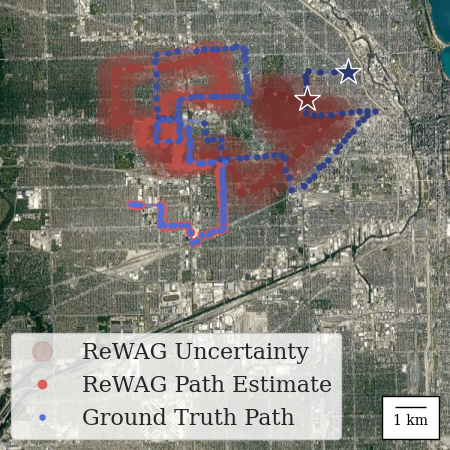}
  \caption{The ground-truth path in Chicago and ReWAG's path estimate, which accurately converges upon the ground-truth. Uncertainty bubble sizes are scaled down by raising to 0.75 to improve interpretability. Increasing brightness of path indicates passing of time. Start marked with stars.}
  \label{fig:path}
  \vskip-0.2in
\end{figure}

\textbf{Experiment details.} The true path the simulated agent travels in Chicago is shown in Fig.~\ref{fig:path} with ReWAG's estimated location at each time step. We ran this experiment with ReWAG and a baseline that uses stage-1 TransGeo \cite{zhu2022transgeo}, a Vi-T approach, for its Siamese network combined with the same particle filter as ReWAG.

\begin{figure}[t!]
  \centering
  \subfigure[Initial distribution supplied to particle filter. True location is over 1 km from initial particle filter estimate. \label{fig:initial_gauss}]{\includegraphics[width=.49\columnwidth]{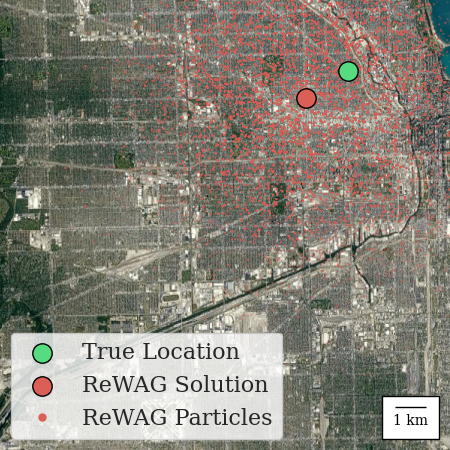}}
  \subfigure[Final particle filter solution and true location. True location is approximately 26 m from final particle filter estimate. \label{fig:zoom_nopart}]{\includegraphics[width=.49\columnwidth]{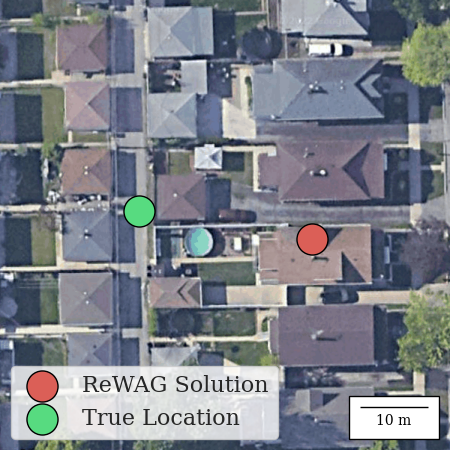}}
  \caption{ReWAG is able to accurately localize an agent to within 26 meters of its true location across nearly 200 km$^2$ of Chicago after being initialized to a Gaussian distribution centered 1.3 km from the true location.}
  \label{fig:bam}
\end{figure}

\begin{figure}[t]
\centering
  \includegraphics[width=0.85\linewidth]{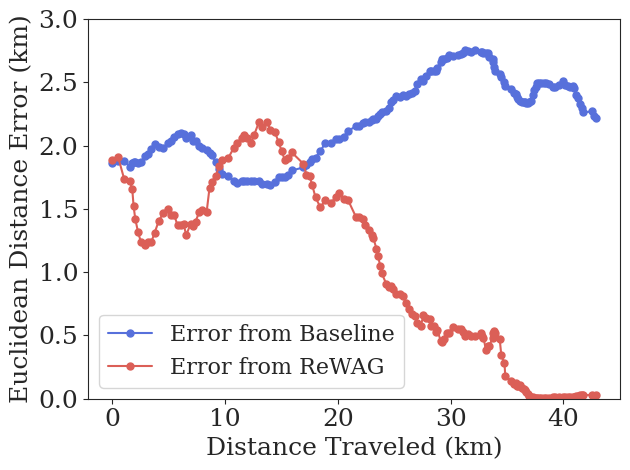}
  \vspace*{-0.2in}
\caption{Particle filter estimation error from ReWAG compared to a TransGeo baseline. ReWAG has lower final and average error.}
  \label{fig:err_conv}
\centering
  \includegraphics[width=0.85\linewidth]{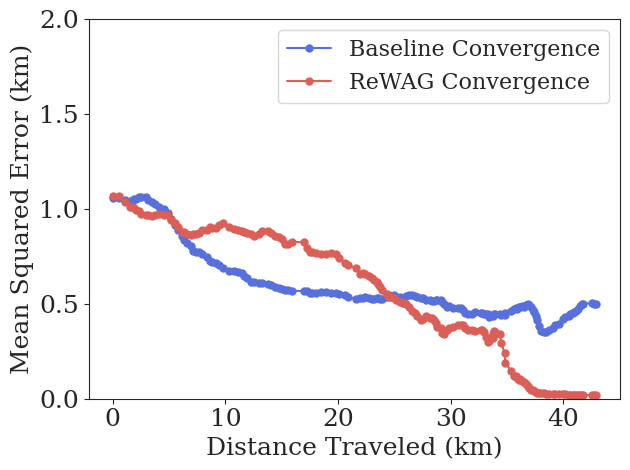}
  \vspace*{-0.2in}
\caption{Particle filter estimation convergence from ReWAG compared to a TransGeo baseline. ReWAG accurately converges, the baseline does not.}
  \label{fig:mse_conv}
\end{figure}

\textbf{Estimation error.} Even with an initialization as far from the agent as Fig.~\ref{fig:initial_gauss}, ReWAG has a final estimation error of 26 m (visible in Fig.~\ref{fig:zoom_nopart}). This estimation error is only 5 meters greater than that which was achieved with 360\textdegree{} ground images in WAG. Fig.~\ref{fig:err_conv} compares the estimation error of ReWAG's particle filter and the TransGeo baseline as the simulated agent moves. The error is the Euclidean distance between the actual location and the weighted average of the particle locations.  Over the duration of the experiment, ReWAG has an average estimation error of 925 m, versus 2.2 km for the baseline. ReWAG achieves a final estimation error of 26 m compared to the baseline of 2.2 km. 

\begin{figure}[t]
\centering
  \includegraphics[width=0.8\linewidth]{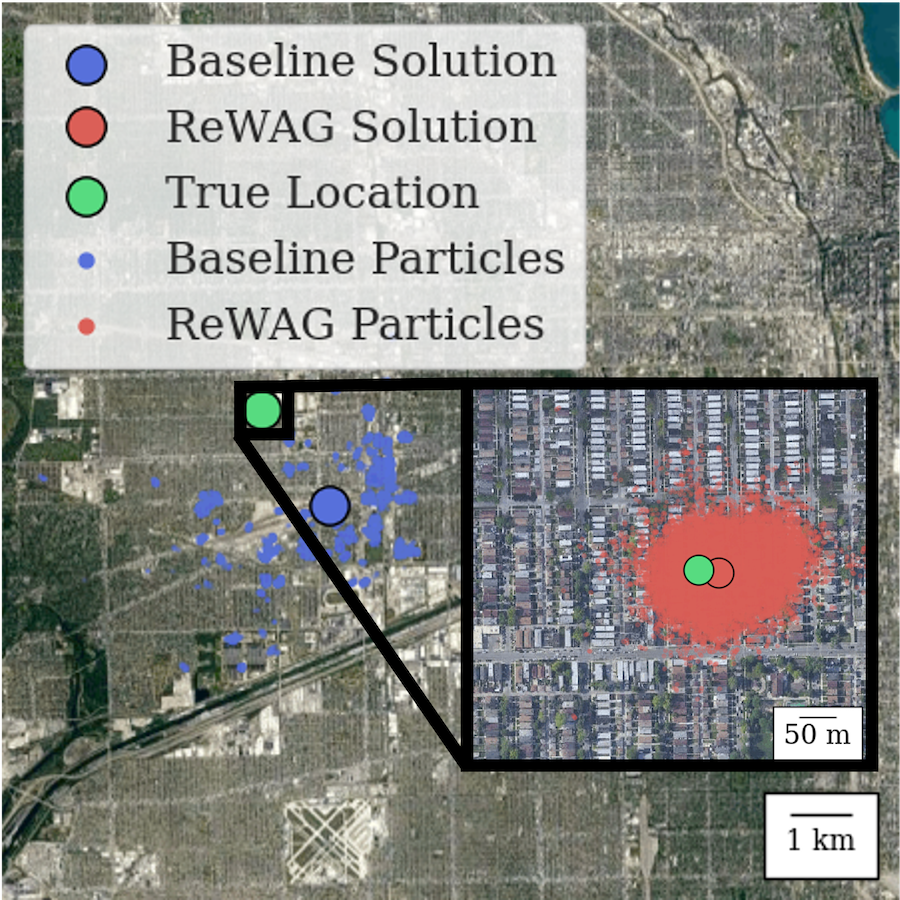}
  \vspace*{-0.1in}
\caption{Final particle filter dispersion with the baseline system and with ReWAG on the Chicago test path (C-1). Baseline does not successfully converge to a location estimate, while ReWAG converges to within 18 m of standard deviation.}
  \label{fig:exp_pf}
  \vspace*{.1in}
\centering
  \includegraphics[width=0.8\linewidth]{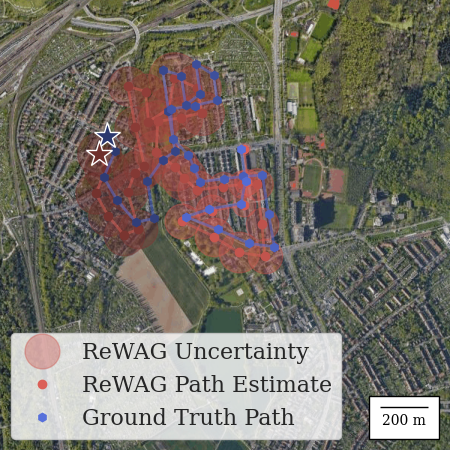}
\vspace*{-0.1in}
\caption{Ground-truth and estimated path in the KITTI test area. Increasing brightness of path indicates passing of time.  Start marked with stars.}
  \label{fig:kitti_path} 
\end{figure}

\textbf{Convergence.} Figure~\ref{fig:mse_conv} compares the system convergence measured as the mean squared error (MSE) of the particle locations at each time step. ReWAG converges to an MSE of less than 60 m (the satellite image size) after 133 filter updates, while the baseline does not reach that level of convergence in the time period tested. Fig.~\ref{fig:exp_pf} shows the particle filter distribution that ReWAG converges to as red dots in the inset of the figure while the baseline terminates with the particle distribution shown in blue dots, which still shows significant estimation error.

\textbf{Ablation.} We performed a small ablation study to determine the benefit of including both heading and location information in the pose-aware embeddings as opposed to only including heading, or including neither as in WAG. We trained a Siamese network with the same architecture, training regime and parameters as ReWAG, with the exception that the base embeddings input to SAFA only had heading appended to them. We also tested limited FOV images with WAG retrained on limited FOV (without heading or position appended). We tested the systems on the C-1 test path and the results are summarized in Table \ref{tab:summary_ablation} We attribute ReWAG without position's performance to the fact that heading alone does not dictate what content is visible in a ground image and hence may be misleading on its own.
\begin{table}[t]
\caption{Ablation}
\vspace*{-.2in}
\label{tab:summary_ablation}
\begin{center}
\begin{tabular}{|cl|c|c|c|}
\hline
\multicolumn{2}{|c|}{\bf Metric and System Type}  & \bf C-1  \\\hline
\multirow{3}{*}{Final Error (m)} & ReWAG & 26\\
& ReWAG without Position & 375 \\
& WAG & 192\\\hline
\multirow{3}{*}{Convergence Time (time steps)} & ReWAG & 133 \\
& ReWAG without Position & - \\
& WAG  & - \\\hline
\end{tabular}
\end{center} \vspace*{-0.1in}
\end{table} 

\setlength{\textfloatsep}{5pt}




\textbf{Multiple path result summary.} Table \ref{tab:summary_results} shows a summary of ReWAG's localization performance compared to the baseline on three test paths in Chicago. These paths are the same test paths that were used in \cite{downes_iros_22}. The first, C-1, is the path discussed previously (Fig.~\ref{fig:path}). The particle filter for C-1 and C-2 was initialized 1.2 km away from the ground truth, and for C-3 was initialized 600 m away from the ground truth due to the challenging urban scenery in this path. On all paths, ReWAG outperforms the baseline in final estimation error, final standard deviation and convergence time. 
\begin{table}[t]
\caption{Comparison of Results on Chicago Test Paths--\\ Baseline: With TransGeo}
\vspace*{-.2in}
\label{tab:summary_results}
\begin{center}
\begin{tabular}{|cl|c|c|c|}
\hline
\multicolumn{2}{|c|}{\bf Metric and System Type}  & \bf C-1  & \bf C-2 & \bf C-3 \\\hline
\multirow{2}{*}{Final Error (m)} & Baseline & 2218 & 2259 & 300\\
& ReWAG & 26 & 16 & 17\\\hline
\multirow{2}{*}{Final Standard Deviation (m)} & Baseline & 500 & 1321 & 169\\
& ReWAG & 18 & 10 & 10\\\hline
\multirow{2}{*}{Convergence Time (time steps)} & Baseline & - & - & -\\
& ReWAG & 133 & 61 & 41\\\hline
\end{tabular}
\end{center} \vspace*{-0.1in}
\end{table} 
\vspace*{-0.2in}
\subsection{Small-scale Test: KITTI}
We also demonstrate ReWAG's localization performance on a KITTI test path in Fig.~\ref{fig:kitti_path}. This experiment demonstrates ReWAG's ability to localize with more accurate initialization information across a smaller search area and its robustness towards localizing on images in a different city than those that it was trained on. We sample images and pose data from the residential ``2011{\_}0{\_}30{\_}drive{\_}0028'' path and reduce the FOV to 90\textdegree{} by cropping the KITTI images. The satellite images of the search area are obtained from the Google Static API; the search area is divided into a grid of 32 $\times$ 32 satellite images at zoom level 20. We initialize the particle filter with a Gaussian distribution approximately 80 m away from the true location, and ReWAG's final estimation error is 12 m after 34 ground image updates.

\section{Conclusion}
ReWAG redesigns the Siamese network-particle filter architecture for increased hardware flexibility, a key component of applying cross-view geolocalization to mobile robotics. Previous works have largely focused on geolocalization with ground cameras of unrealistic FOVs for existing robotics platforms. This work uses all available information to generate Siamese network embeddings and accurately reweight particles, hence enabling faster and more accurate particle filter convergence with limited FOV cameras. 

ReWAG accurately localizes across several hundred square kilometers of Chicago, maintaining the same level of accuracy that WAG \cite{downes_iros_22} previously demonstrated with panoramic ground cameras. ReWAG maintains the same benefits of WAG in terms of reducing the size of the satellite image database required. Additionally, it has lower final error and faster convergence than the TransGeo baseline in the Chicago test area, which we attribute to its pose-awareness. It also demonstrates its ability to generalize to a city it was not trained on by successfully converging with only 12 m error in the KITTI test path.

In the short term, future work on this topic includes improving computational speed, testing on real-time physical platforms, and further restriction of the FOV. Domain shift also remains an open challenge in the field--- localizing in areas with different appearances than training image pairs, on satellite images that were taken in different seasons than the ground images, and in rural areas without as many identifiable landmarks as residential or urban areas.

In summary, ReWAG lifts the heavy hardware requirements that were inherent in other cross-view geolocalization systems and enables accurate localization using a camera with a narrow FOV. It is a step forward in making cross-view geolocalization a widespread, cross-platform tool. 

\balance
\bibliographystyle{IEEEtran}
\bibliography{root}

\end{document}